\documentclass[conference]{IEEEtran}
\IEEEoverridecommandlockouts
\usepackage{cite}

\usepackage{amsmath,amssymb,amsfonts}
\usepackage{algorithmic}
\usepackage{graphicx}   
\usepackage{textcomp}
\usepackage{xcolor}
\usepackage[hidelinks]{hyperref}
\usepackage{stfloats}
\usepackage{booktabs, tabularx}
\usepackage{rotating}
\usepackage{longtable}
\usepackage{soul}
\usepackage{subfigure}
\usepackage{url}
\usepackage{float}
\usepackage{array}
\usepackage{multirow}
\usepackage{ragged2e}
\usepackage{tabularray}

\usepackage [english]{babel}
\usepackage [autostyle, english = american]{csquotes}
\MakeOuterQuote{"}

\def\BibTeX{{\rm B\kern-.05em{\sc i\kern-.025em b}\kern-.08em
    T\kern-.1667em\lower.7ex\hbox{E}\kern-.125emX}}

\newcommand{\refappendix}[1]{\hyperref[#1]{Appendix~\ref*{#1}}}

\begin{document}

\title{Risk Assessment of an Autonomous Underwater Snake Robot in Confined Operations\\

\author{Abdelrahman Sayed Sayed
     
		\thanks{Abdelrahman Sayed Sayed is with Department of Marine Technology,
        Norwegian University of Science and Technology (NTNU), Otto Nielsens veg 10, 7491 Trondheim, Norway; Université de Toulon, Toulon, France.
		{\tt\small abdelrahman.s.s.e.ibrahim@ntnu.no,
            abdelrahman-ibrahim@etud.univ-tln.fr
            }}%
			
	}

}

\maketitle
%comment this to remove page numbers
% \thispagestyle{plain}
% \pagestyle{plain}

\begin{abstract}

The growing interest in ocean discovery imposes a need for inspection and intervention in confined and demanding environments. Eely's slender shape, in addition to its ability to change its body configurations, makes articulated underwater robots an adequate option for such environments. However, operation of Eely in such environments imposes demanding requirements on the system, as it must deal with uncertain and unstructured environments, extreme environmental conditions, and reduced navigational capabilities. 
This paper proposes a Bayesian approach to assess the risks of losing Eely during two mission scenarios. The goal of this work is to improve Eely's performance and the likelihood of mission success. Sensitivity analysis results are presented in order to demonstrate the causes having the highest impact on losing Eely.
\end{abstract}


\section{Introduction}
\label{sect:1}

The growing interest in ocean discovery imposes a need for inspection and intervention in confined and demanding environments. Underwater confined environments, such as shipwrecks and sunken caves, present unique challenges for exploration. These environments often have limited access points and tight spaces, making it difficult for divers and underwater vehicles to enter and maneuver \cite{intro1}. Autonomous underwater vehicles (AUVs) are considered to be efficient sensor-carrying platforms for seabed mapping and monitoring \cite{intro2}. The likelihood of the underwater vehicles being lost while performing these missions can under certain circumstances be high, and some AUVs have been lost during missions due to technical failures \cite{intro6}.\\

\begin{figure}[htbp]
\centerline{\includegraphics[width=6cm,height=5cm]{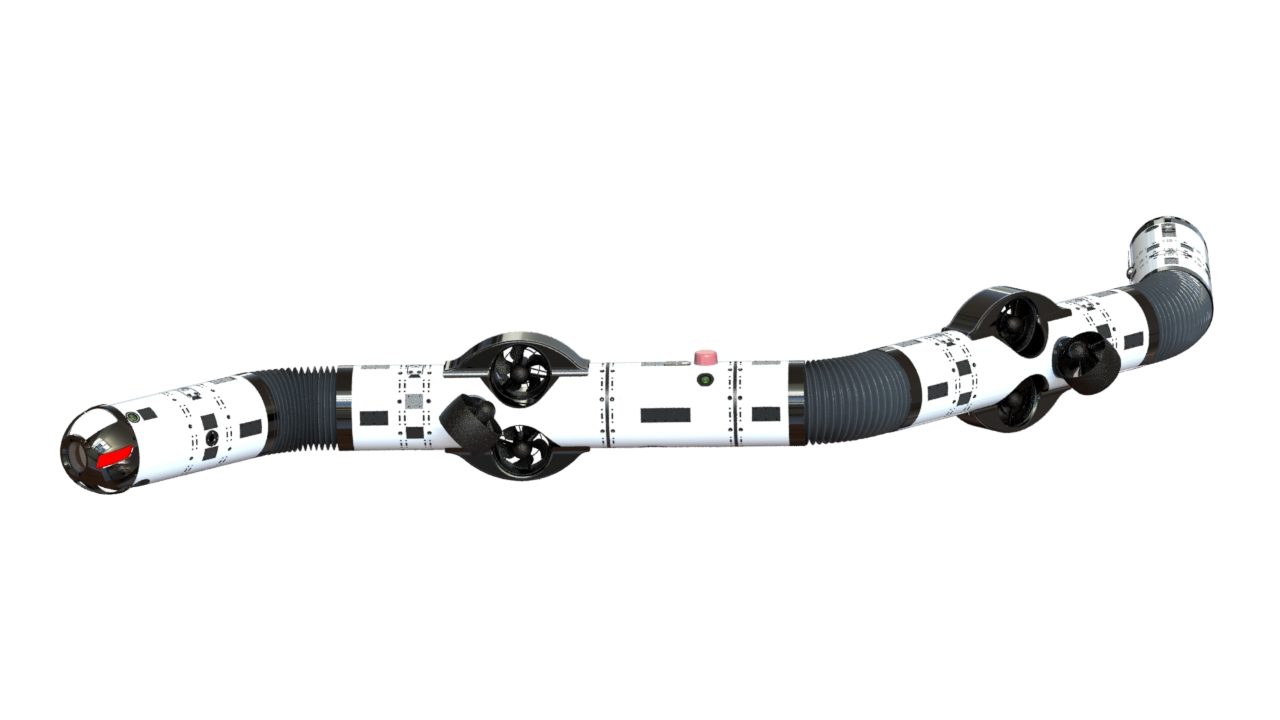}}
\caption{Triple joint snake robot from Eelume}
\label{fig:Eely}
\end{figure}

Underwater snake robots, like Eely which is a snake robot from Eelume \cite{eelume,eelume2} is a promising option for these environments due to their slender shape and ability to change body configurations. Eely's articulated structure shown in Figure \ref{fig:Eely} combines the advantages of several types of underwater vehicles, as it has the range of AUVs, the ability to access challenging areas like small Remotely operated underwater vehicles (ROVs), and the intervention capabilities of ROVs \cite{intro7}. Thereby, Eely covers a broad range of operational scenarios as the vehicle can be configured to follow a torpedo-shaped AUV for platforming missions requiring the robot to map a big area \cite{intro8} or to vary the joint's modules to follow snake configuration to map confined or steep environments which can not be achieved by a normal AUV. However, operation of Eely in such environments imposes demanding requirements on the system, as it must deal with uncertain and unstructured environments, extreme environmental conditions, and reduced navigational capabilities \cite{intro9}.

Bayesian networks (BNs) have been widely used to assess the risks associated with Autonomous Underwater Vehicles (AUVs) in operational scenarios, as demonstrated in previous studies \cite{ing1,hazid2,rmodel4,rmodel5}. However, there has been no prior work that has specifically addressed the risks associated with modular underwater snake robots operating in confined environments.\\

The main scientific contribution in this paper, we present a novel approach to risk assessment for Eely, an underwater snake robot, in confined environments operations. Our aim is to improve Eely's performance and increase the likelihood of mission success. Relevant data are collected to perform a quantitative risk analysis to develop a Bayesian model that considers almost the entire system of Eely, thus avoiding potential risks. We demonstrate that our Bayesian model can be extended to a Decision network (DN), enabling Eely to adapt its behavior autonomously and maximize mission utility \cite{intro10}. Although our model was developed specifically for Eely, it can be transferred to the operations of other AUVs. To the best of our knowledge, the topic of risk associated with modular underwater snake robots has not been previously explored in existing research. Therefore, our study presents a state of the art contribution to the field of risk assessment of underwater vehicles.\\

The paper is organized as follows: Section~\ref{sect:2} presents the background on risk and Bayesian networks. In Section~\ref{sect:3}, the application of the proposed BN risk model for the two case scenarios, including data collection for parameters for these models, dynamic simulation, and sensitivity analysis to identify the causes with the most significant influence on losing Eely. In Section~\ref{sect:4} 
the steps to extend the BN to a DN for autonomous risk-based decision making are presented. This is followed by Section~\ref{sect:5}, which gives a brief discussion of the results. Finally, Section~\ref{sect:6} concludes this paper and presents suggestions for future work.

\section{Risk Modeling Using BNs}
\label{sect:2}

\subsection{Definition of Risk}

The definition of risk related to a hazardous event $e_i$ \cite{rmodel1} can be represented by the following relation:

\begin{equation}
    r = \{e_{i},c_{i},q\}|k
\label{Eq:BN1}
\end{equation}

where $c_{i}$ is the consequence of $e_{i}$, $q$ is the measure of involved uncertainty, and $k$ is the background knowledge for determining $e_{i}$, $c_{i}$ and $q$. This is the most commonly used definition for Bayesian risk modeling. By accounting for previous knowledge about the operational conditions and mission scenarios, events with low background knowledge would not have a strong effect when making a decision in contrast to events with high background knowledge. 
According to \cite{rmodel2}, risk assessment is an overall process including risk identification, risk analysis, and risk evaluation. Risk identification identifies and illustrates the possible risks with respect to the mission objective. A common approach for risk identification is to identify the known hazards as a source[s] of prospective harm. These hazards can be later analyzed in the risk analysis step, where the known possible events and their expected outcomes are modeled. Therefore, risk identification is an essential step towards developing a control system with risk management and decision-making capabilities for Eely which is known as supervisory risk control \cite{rmodel3}. There are many of risk identification methods, such as hazard identification (HAZID) and preliminary hazard analysis (PHA). In order to have effective supervisory risk control, the most significant mission and operation hazards should be identified and combined into a risk model to aid in the system's decision-making.

\subsection{Bayesian Networks}

A Bayesian network is a graphical model used to address problems involving uncertainty \cite{rmodel4,rmodel5}. It describes the dependencies between random variables in a directed acyclic graph, where nodes stand for random variables and directed arcs between nodes signify conditional dependencies between them. The Bayesian network, which is based on probability theory \cite{rmodel6}, allows for two-way reasoning by handling both causal reasoning, which derives posterior probabilities from prior probabilities, and diagnostic reasoning, which uses a formula to derive prior probabilities from posterior probabilities. The traditional static Bayesian network (SBN) has limitations when it comes to evaluating variables that change over time, but it is useful for analyzing and forecasting data at a specific time. The dynamic Bayesian network was created to overcome this drawback. The dynamic Bayesian network (DBN) incorporates methods that take into account the relationship between moments in time, known as the state transition probability, in contrast to SBN \cite{rmodel7}. By considering this relationship, the network can learn to effectively use changes in values over time, producing better results.

\section{Construction of Bayesian Risk model}
\label{sect:3}

In the context of risk assessment, Definition (\ref{Eq:BN1}) may imply a Bayesian approach to probabilities, and hence the use of BN is a suitable approach to estimate the probability of losing Eely during different operational scenarios. HAZID was performed for the two case scenarios: i. Seabed mapping and ii. Confined environments operations in the form of a preliminary hazard analysis (PHA) prior to the development of the Bayesian risk model. Using the five-step method described in \cite{intro10}, the Bayesian risk model is developed.

\subsection{HAZID}

HAZID aims to identify the hazards related to the operation of Eely for the two proposed case scenarios. Detectability is defined as Eely's ability to detect and monitor a specific hazard during its operation, where a high degree of detectability suggests a strong background knowledge, while a low degree of detectability suggests a weak background knowledge. The collection of hazards was mainly based on \cite{hazid1} as it offered a comprehensive compilation of potential risks and served as the foundation for the identification procedure, along with our own judgment and similar systems \cite{intro10}. In addition to, previous field trials with Eely in Trondheim Fjord \cite{hazid4} and previous experiments in karstic exploration \cite{hazid5,hazid6,hazid7} were also taken into account.\\

Table \ref{Table:Criteria}, adopted from \cite{intro10} which was used for hazards in the context of AUV operation under ice, the same criteria was selected and modified based on hazards related to Eely's operation. Then, these hazards were assigned possible consequences depending on each case scenario, and the most noticeable hazards were identified. The risk priority number ($rpn$) for each hazard is obtained from the categories shown in Table \ref{Table:Criteria}. The $rpn$ is the product of the frequency rating, the consequence rating, and the detectability rating. To assess the risk, the worst-case scenario is assumed which is losing Eely. Note that this loss may be temporary, as in the case of seabed mapping and deploying a vessel may recover Eely \cite{hazid2}, or it can be a permanent loss as in the case of underwater caves. 
Tables \ref{Table:PHA seabed} - \ref{Table:PHA confined} in the Appendix lists the hazards identified for the two case scenarios presented in the study. The assessment sheets design in the tables is based on the PHA sheet from \cite{hazid3}.

\begin{table}[hbt!] %To Fix the table position
\centering
\caption{Criteria used to rate Frequency, Consequences and Detectability of events}
\begin{tblr}{
  width = \linewidth,
  colspec = {Q[130]Q[75]Q[750]},
  cell{2}{1} = {c=3}{0.939\linewidth},
  cell{6}{1} = {c=3}{0.939\linewidth},
  cell{10}{1} = {c=3}{0.939\linewidth},
  hlines,
  vlines,
}
\textbf{Rating} & \textbf{Score} & \textbf{Description}\\
\textbf{Frequency}     &       & \\
Low           & 1     & The event may occur less than once per mission\\
Medium        & 2     & The event will be encountered, on average, once per mission\\
High          & 3     & The event will be encountered several times per mission\\
\textbf{Consequences}  &       & \\
Low/None      & 1     & The event will have no negligible influence on the mission with respect to damage to Eely or loss of mission data                 \\
Medium        & 2     & The event may lead to damages or delays that will minorly reduce the time available for the mission or affect the data collection \\
High          & 3     & The event may lead to loss of Eely, early abortion of the mission, or significant loss of scientific data\\
\textbf{Detectability} &       & \\
Low/None      & 3     & Eely is not able to detect or assess the hazardous event during the operation\\
Medium        & 2     & Eely may infer information about the hazardous event. However, the inference will be associated with high uncertainty\\
High          & 1     & Eely may collect and infer information about the hazardous event with high certainty        
\end{tblr}
% Adopted from \cite{D1}&  & 
\label{Table:Criteria}
\end{table}

\subsection{Failure Probabilities}

Some of the issues related to conducting a quantitative risk assessment are obtaining the failure data of the sensors and components of the robot. Since Eely robot is in the early stages of experiments, the availability of its failure data is scarce. Different data sources were used, which include literature and technical specifications, etc. For the non-available data, assumptions based on overall knowledge were applied, which provides rough estimates for the scenarios.\\

The failure probabilities per year are given in Table \ref{Table:failures}, which are used in the BN nodes. However, it should  be noted that failure probabilities for some parts of the robot’s sensors were taken as for the general parts and not for a product of a specific brand from a specific company. For example, the failure probability for the Underwater hyperspectral imager (UHI) sensor was taken as the same value for the general camera as the UHI is simply a more advanced camera as it combines a push-broom hyperspectral imager with an external light source \cite{UHI1}. The obtained data which is available is used as an initial attempt for quantitative risk assessment, but more proper assumptions and precise data could be extracted in future trials and for system suppress analysing own design.

\begin{table}[hbt!]
\centering
\caption{FAILURE PROBABILITIES}
\begin{tblr}{
  width = \linewidth,
  colspec = {Q[408]Q[285]Q[229]},
  vline{2-3} = {-}{},
  hline{1-2,11} = {-}{},
}
Sensor/Item               & Failure Probability & Source            \\
Joint module actuators    & 0.125               &  \cite{intro1}                 \\
Thruster module actuators & 0.1                 &  \cite{failure8}                 \\
DVL sensor                & 0.1                 &  \cite{failure5}                 \\
IMU sensor                & 0.01                &  \cite{failure6}               \\
UHI sensor                & 0.01                & \cite{failure1}    \\
Cameras                   & 0.01                & \cite{failure1}    \\
LED lights                & 0.02                & \cite{failure3}    \\
Leakage sensor            & 0.05                & \cite{failure2}    \\
Batteries                 & 0.000001            & \cite{failure4} 
\end{tblr}
\label{Table:failures}
\end{table}

\subsection{Conditional Probability Tables}

The conditional probability tables (CPT) \ref{Table:cpt}-\ref{Table:cpt loss of Eely} are quantified based on the PHA. Table \ref{Table:cpt} shows the CPT for the failure of the propulsion system node. This table shows how critical is it in case of the failure of thruster module actuators as shown in red in any case of failure of thruster module actuators the probability of losing Eely is true as in confined environments the robot can't use its neutral buoyancy to float back to the surface. This is further illustrated in the sensitivity analysis subsection. Due to the limit on the number of pages, the CPTs for other nodes are omitted from this version of the paper. Moreover, thrusters were one of the most susceptible components to failure as Eely operators at the Applied Underwater Robotics Laboratory (AUR Lab) have communicated that they have lost on average one thruster per year.

\begin{table}[hbt!]
\centering
\caption{CPT for failure of propulsion system}
\begin{tblr}{
  width = \linewidth,
  colspec = {Q[498]Q[48]Q[48]Q[54]Q[54]Q[56]Q[56]Q[67]Q[38]},
  cells = {c},
  cell{1}{2} = {c=4}{0.204\linewidth},
  cell{1}{6} = {c=4}{0.217\linewidth},
  cell{2}{2} = {c=2}{0.096\linewidth},
  cell{2}{4} = {c=2}{0.108\linewidth},
  cell{2}{6} = {c=2}{0.112\linewidth},
  cell{2}{8} = {c=2}{0.105\linewidth},
  cell{4}{2} = {red},
  cell{4}{3} = {red},
  cell{4}{4} = {red},
  cell{4}{5} = {red},
  cell{4}{9} = {red},
  cell{5}{2} = {red},
  cell{5}{3} = {red},
  cell{5}{4} = {red},
  cell{5}{5} = {red},
  cell{5}{9} = {red},
  hlines,
  vlines,
}
Failure of thruster module actuators & TRUE &   &       &   & FALSE &     &       &   \\
Failure of joint module actuators    & TRUE &   & FALSE &   & TRUE  &     & FALSE &   \\
Environmental complexity             & T    & F & T     & F & T     & F   & T     & F \\
TRUE                                 & 1    & 1 & 1     & 1 & 0.3   & 0.1 & 0.35   & 0 \\
FALSE                                & 0    & 0 & 0     & 0 & 0.7   & 0.9 & 0.65   & 1 
\end{tblr}
\label{Table:cpt}
\end{table}

\begin{table}[hbt!]
\centering
\caption{CPT for Environmental Complexity}
\begin{tblr}{
  width = \linewidth,
  colspec = {Q[200]Q[48]Q[48]Q[54]Q[54]Q[56]Q[56]Q[67]Q[38]},
  cells = {c},
  cell{1}{2} = {c=4}{0.204\linewidth},
  cell{1}{6} = {c=4}{0.217\linewidth},
  cell{2}{2} = {c=2}{0.096\linewidth},
  cell{2}{4} = {c=2}{0.108\linewidth},
  cell{2}{6} = {c=2}{0.112\linewidth},
  cell{2}{8} = {c=2}{0.105\linewidth},
  cell{4}{2} = {},
  cell{4}{3} = {},
  cell{4}{4} = {},
  cell{4}{5} = {},
  cell{4}{9} = {},
  cell{5}{2} = {},
  cell{5}{3} = {},
  cell{5}{4} = {},
  cell{5}{5} = {},
  cell{5}{9} = {},
  hlines,
  vlines,
}
Ocean current & TRUE &   &       &   & FALSE &     &       &   \\
Dusty sediments    & TRUE &   & FALSE &   & TRUE  &     & FALSE &   \\
Absence of natural light             & T    & F & T     & F & T     & F   & T     & F \\
TRUE                                 & 0.95    & 0.9 & 0.6     & 0.4 & 0.7   & 0.6 & 0.15   & 0.01 \\
FALSE                                & 0.05    & 0.1 & 0.4     & 0.6 & 0.3   & 0.4 & 0.85   & 0.99 
\end{tblr}
\label{Table:cpt environmental}
\end{table}

\begin{table}[hbt!]
\centering
\caption{CPT for Remote Control}
\begin{tblr}{
  width = \linewidth,
  colspec = {Q[200]Q[48]Q[48]Q[54]Q[54]Q[56]Q[56]Q[67]Q[38]},
  cells = {c},
  cell{1}{2} = {c=4}{0.204\linewidth},
  cell{1}{6} = {c=4}{0.217\linewidth},
  cell{2}{2} = {c=2}{0.096\linewidth},
  cell{2}{4} = {c=2}{0.108\linewidth},
  cell{2}{6} = {c=2}{0.112\linewidth},
  cell{2}{8} = {c=2}{0.105\linewidth},
  cell{4}{2} = {},
  cell{4}{3} = {},
  cell{4}{4} = {},
  cell{4}{5} = {},
  cell{4}{9} = {},
  cell{5}{2} = {},
  cell{5}{3} = {},
  cell{5}{4} = {},
  cell{5}{5} = {},
  cell{5}{9} = {},
  hlines,
  vlines,
}
Failure of communication system & TRUE &   &       &   & FALSE &     &       &   \\
Failure of operator intervention    & TRUE &   & FALSE &   & TRUE  &     & FALSE &   \\
Environmental complexity             & T    & F & T     & F & T     & F   & T     & F \\
TRUE                                 & 0.4    & 0.3 & 0.3     & 0.2 & 0.1   & 0.01 & 0.05   & 0.01 \\
FALSE                                & 0.6    & 0.7 & 0.7     & 0.8 & 0.9   & 0.99 & 0.95   & 0.99 
\end{tblr}
\label{Table:cpt remote ctrl}
\end{table}

\begin{table}[hbt!]
\centering
\caption{CPT for Loss of Eely}
\begin{tblr}{
  width = \linewidth,
  colspec = {Q[465]Q[108]Q[121]Q[108]Q[121]},
  cells = {c},
  cell{1}{2} = {c=2}{0.228\linewidth},
  cell{1}{4} = {c=2}{0.228\linewidth},
  hlines,
  vlines,
}
Failure of autonomous control & TRUE &       & FALSE &       \\
Failure of remote control     & TRUE & FALSE & TRUE  & FALSE \\
TRUE                          & 1    & 0.75  & 0.1   & 0     \\
FALSE                         & 0    & 0.25  & 0.9   & 1     
\end{tblr}
\label{Table:cpt loss of Eely}
\end{table}

\subsection{Launching the BN Model}

After gathering the failure probabilities data, the BN model is constructed using the software GeNIe 4.0, which allows interactive model building and learning, developed by Bayes Fusion company \cite{genie1}.

\subsection{Dynamic Simulation}

\begin{figure*}[htbp]
\centering
\includegraphics[width=\textwidth,height=10cm]{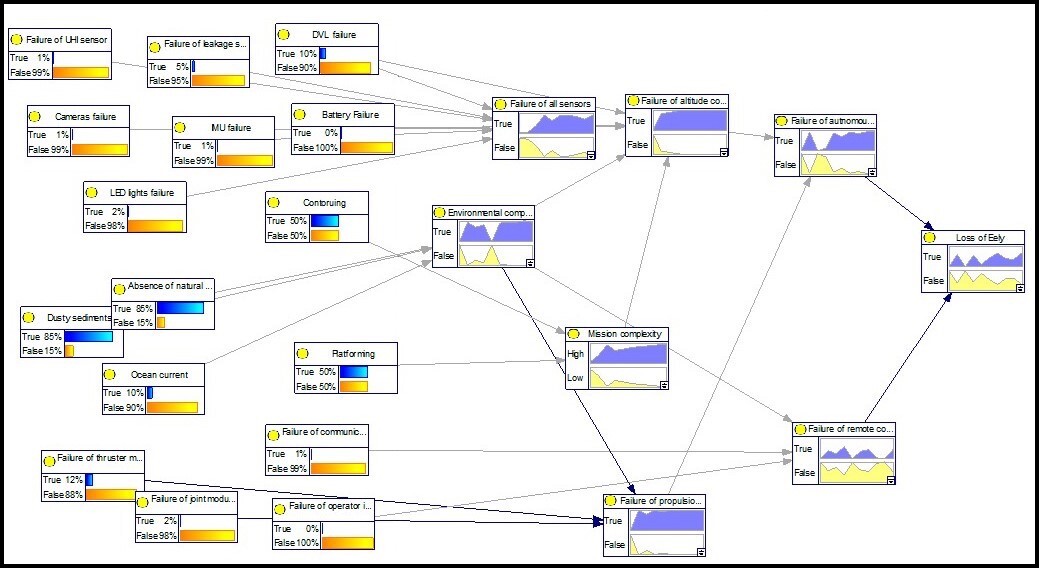}
\caption{DBN for Losing Eely during Confined Environments Operations}
\label{fig:DBN_confined}
\end{figure*}

\begin{figure*}[htbp]
\centering
\begin{minipage}{.45\textwidth}
\centering
\subfigure[Environmental Complexity]{\includegraphics[width=8cm,height=1cm]{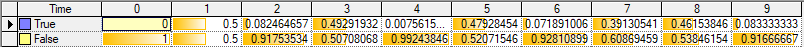}}
\subfigure[Mission Complexity]{\includegraphics[width=8cm,height=1cm]{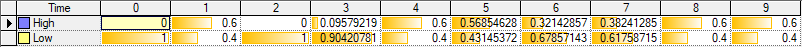}}
\subfigure[Failure of Propulsion System]{\includegraphics[width=8cm,height=1cm]{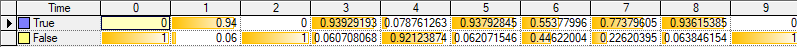}}
\subfigure[Failure of Remote Control]{\includegraphics[width=8cm,height=1cm]{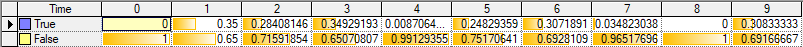}}
\subfigure[Loss of Eely]{\includegraphics[width=8cm,height=1cm]{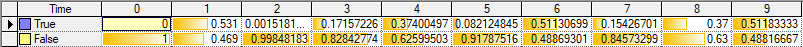}}
\caption{Dynamic Simulation Results for Seabed Mapping Operations}
\label{fig:DBN seabed}
\end{minipage}%
\hfill
\begin{minipage}{.45\textwidth}
\centering
\subfigure[Environmental Complexity]{\includegraphics[width=8cm,height=1cm]{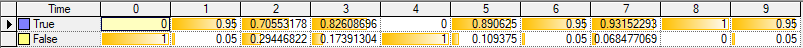}}
\subfigure[Mission Complexity]{\includegraphics[width=8cm,height=1cm]{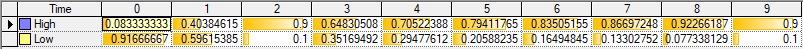}}
\subfigure[Failure of Propulsion System]{\includegraphics[width=8cm,height=1cm]{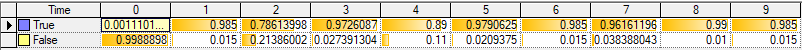}}
\subfigure[Failure of Remote Control]{\includegraphics[width=8cm,height=1cm]{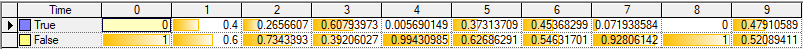}}
\subfigure[Loss of Eely]{\includegraphics[width=8cm,height=1cm]{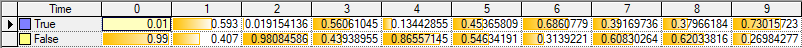}}
\caption{Dynamic Simulation Results for Confined Environments Operations}
\label{fig:DBN confined}
\end{minipage}
\end{figure*}

Figure \ref{fig:DBN_confined} shows one of the developed DBNs for the case of confined environments operations. The simulation results of the DBNs for the two case scenarios can be observed in Figures \ref{fig:DBN seabed} - \ref{fig:DBN confined}. The results indicate that the DBN can take into account changes over time and the likelihood of events occurring. In this regard, the probability of losing Eely in confined environments is higher than in seabed mapping due to various factors such as extreme pressures on the thrusters, poor communication with the vehicle, high mission complexity, and the complexity of the environment.

\subsection{Sensitivity Analysis}

\begin{figure*}[htbp]
\centering
\begin{minipage}{0.48\textwidth}
\centering
\includegraphics[width=\linewidth]{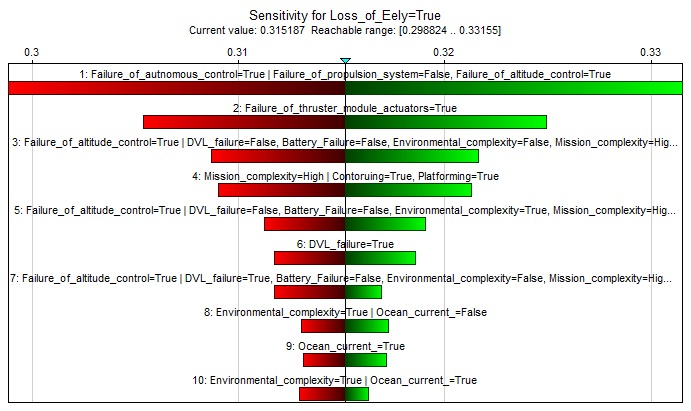}
\caption{Sensitivity tornado diagram for losing Eely during Seabed Mapping Operations}
\label{fig:SA_Seabed}
\end{minipage}\hfill
\begin{minipage}{0.48\textwidth}
\centering
\includegraphics[width=\linewidth]{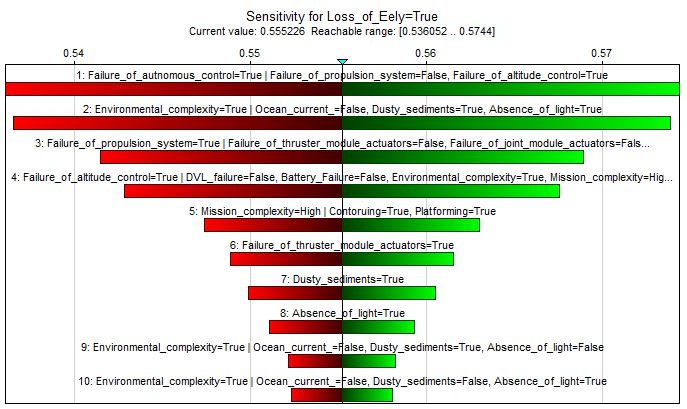}
\caption{Sensitivity tornado diagram for losing Eely during Confined Environments Operations}
\label{fig:SA_Tunnel}
\end{minipage}
\end{figure*}

The goal of sensitivity analysis is to identify the causes that have the most significant impact on the probability of losing Eely, and to limit these causes by introducing risk reduction measures. It also works as an indicator for adding more constraints and efforts for data collection. Decreasing the uncertainty of a cause that has little or no influence on the probability of losing Eely will result in a negligible change in the overall uncertainty value, making it of little importance.
In our case, we aim to identify the top factors that influence the probability of losing Eely. Figures \ref{fig:SA_Seabed} - \ref{fig:SA_Tunnel} show the sensitivity analysis results for losing Eely for the two case scenarios.\\

In the first case of seabed mapping operations shown in Figure \ref{fig:SA_Seabed}, it can be seen that the probability of losing Eely is most sensitive to the failure of autonomous control. Other main factors are comprised of failure of thruster module, failure of altitude control, mission complexity, and DVL failure.
The second case of sensitivity analysis is performed for confined environment operations and is shown in Figure \ref{fig:SA_Tunnel}. It can be seen that the failure of autonomous control is the most sensitive factor for losing Eely as well. Environmental complexity, failure of propulsion system, failure of altitude control, and mission complexity account for the remaining factors to which the loss of Eely is sensitive.

\subsection{Pros and Cons of the BN Model}

The model looks upon almost Eely’s entire system to avoid potential risks. Despite the fact that the model was built specifically for Eely, it can be transferred to the operations of other autonomous underwater vehicles. 
On the other hand, the model has some limitations, such as not including all of the possibilities with details which can cause the loss of Eely and not completely extending some nodes. For instance, nodes such as failure of communication system and failure of operator intervention, the latter being a non-technical node, could be further developed, and other non-trivial factors could have been included in the model. For such expansion, it would require searching for more data including human and organizational factors and their influence on mission risk, which would take a significant amount of time and resources, having the current model seem to be adequate for the two investigated case scenarios. Moreover, DBN is time dependent where the state of a variable at one time depends on its previous states and the states of other variables and it can be computationally expensive to compute more time steps, and GeNIe is only limited to 1000 time step \cite{genie1}.

\section{Risk-Based Decision Making}
\label{sect:4}

Bayesian probability theory is used by both DNs and DBNs to represent uncertain relationships between variables. The main difference is that DNs do not take into account time dependence and provide a snapshot of the system at a specific moment. DNs are employed to simulate the relationship between choices and results when making decisions. In contrast, DBNs can also include decision-making only as part of a dynamic system that changes over time.\\ 

For decision-making, the novel approach presented in \cite{intro10} can be followed, in which the HAZID results of Eely's operations can be used as a basis for constructing the BN model. The BN model can then be extended to a DN to autonomously adapt Eely’s behavior with decision nodes, i.e., by adjusting the altitude set point, speed set point, and control strategy \cite{intro8}. In our case, the maximum joint angle for dynamically changing the robot’s shape can be adjusted based on its belief about the current state of the risk.

\subsection{Decision Nodes}

The decision nodes that should be directly controlled by Eely are: altitude/depth set point, speed set point, the control strategy. In the articulated snake robot case, we can add one more node, “changing shape” based on mission category and environmental constraints. Therefore, our decision nodes could be as follows:

\begin{itemize}
    \item D1 = $a_{s}$ as the altitude set point
    \item D2 = $v_{s}$ as the speed set point
    \item D3 = $c_{s}$ as the control strategy
    \item D4 = $s_{c}$ as the desired/optimum shape configuration
\end{itemize}

\subsection{Online Reasoning}

The DN continuously updates based on the sensory and temporal contexts. It is also important to include dwelling time based on the task to avoid rapid switching of states generated from sensor noise or in transient situations in-between state transitions. This can also help to avoid sensor outliers that bypassed the filtering process.

\subsection{Hard Coding Safety/contingency Handling}

This will impose predefined safety protocols that override the suggested control actions from the DN and perform strictly defined actions, such as Eely's preexisting safety functions, like collision avoidance.

\section{Discussion}
\label{sect:5}

From the results of the DBN and senstivitity analysis, improving the robustness of the autonomous control part would significantly decrease the risk of losing Eely robot during different operations. Regarding the seabed mapping operations, from the sensitivity tornado graph in Figure \ref{fig:SA_Seabed} shows that improving the thruster module actuators and altitude control systems would also substantially reduce the risk of losing Eely. As for confined space operations, the sensitivity tornado graph in Figure \ref{fig:SA_Tunnel} indicates that, except for environmental and mission complexity, improving the robustness of the thruster module actuators, propulsion system, and altitude control system would also significantly reduce the risk of losing Eely. The uncertainties associated with confined environments environmental complexity are reflected in the DBN in Figure \ref{fig:DBN_confined} as they affect critical nodes, and in Tables \ref{Table:PHA seabed} - \ref{Table:PHA confined} as they are the main reason for a high $rpn$, mainly for DVL failure and Controller failure.

\section{CONCLUSION AND FUTURE WORK}
\label{sect:6}

In this paper, a BN model is developed for risk assessment of autonomous operations for Eely underwater snake robot. The model is populated with data to perform a specific quantitative probabilistic estimation of the loss of Eely during two challenging mission scenarios. The data is based on literature, PHA, and our
own judgement based on similar systems. The results from the dynamic simulation and sensitivity analysis show that the highest risk of losing Eely was during confined environments operations, which is related to many factors, but the highest of them is the uncertainty of these environments compared to the other case scenario. 
In the future, this work can be build upon to include more mission scenarios. Also, implementing a behavior tree can allow for more complex mission scenarios and improve the modularity of Eely control systems \cite{conc1} during different mission scenarios.

\section*{Acknowledgment}
I would like to thank my master thesis supervisors from NTNU, Ingrid B. Utne and Asgeir J. Sørensen, for their valuable inputs to this paper. Co-funded by the Erasmus Mundus Joint Master's Degree in Marine and maritime Intelligent Robotics (MIR) a Erasmus+ Programme of the European Union.

\bibliographystyle{IEEEtran}
\bibliography{REF}

% Appendix section

% \newpage
% \vspace{450pt}
\section*{Appendix}
\onecolumn
% \section*{Appendix}
%///////////////////////
% \usepackage{tabularray}
\begin{table}
\centering
\caption{Relevant hazards for seabed mapping operations of Eely}
\resizebox{\linewidth}{!}{%
\begin{tblr}{
  width = \linewidth,
  colspec = {Q[70]Q[175]Q[250]Q[250]Q[50]Q[50]Q[50]Q[40]},
  row{1} = {c},
  row{2} = {c},
  cell{1}{1} = {r=2}{},
  cell{1}{2} = {r=2}{},
  cell{1}{3} = {r=2}{},
  cell{1}{4} = {r=2}{},
  cell{1}{5} = {c=4}{0.161\linewidth},
  cell{3}{1} = {r=2}{},
  cell{5}{1} = {r=2}{},
  cell{8}{1} = {r=3}{},
  cell{11}{1} = {r=2}{},
  cell{13}{1} = {r=2}{},
  cell{15}{1} = {r=2}{},
  vlines,
  hline{1,3,5,7-8,11,13,15,17} = {-}{},
  hline{2} = {5-8}{},
  hline{4,6,9-10,12,14,16} = {2-8}{},
}
\textbf{Hazard } & \textbf{Hazardous Event } & \textbf{Cause } & \textbf{Consequence} & \textbf{\textbf{Risk}} &  &  & \\
 &  &  &  & Freq. & Conseq. & Detect. & Prod. (rpn)\\
Strong currents. & Currents are too strong for Eely to handle. & Currents are not accounted for in the mission plan. & Reduced mission duration. Reduced maneuverability in different configurations for Eely. & 1 Low & 2 Med & 1 High & 2\\
 & Strong currents lead Eely to hit the seabed. & Eely is doing contouring for a steep area, wall. & Damage to Eely or one of its modules. Loss of Eely. & 1 Low & 3 High & 2 Med & 6\\
Contact with seabed. & Eely collides with seabed & Failure to detect the seabed. Failure to follow the safe distance between Eely and the seabed. & Damage to Eely or one of its modules. Loss of Eely. & 1 Low & 3 High & 1 High & 3\\
 & Eely is stuck in soft sediment and algae in the seabed. & Failure to detect the seabed. Failure to detect algae or OOI. Failure to follow the safe distance between Eely and the seabed. & Damage to Eely or one of its modules. Loss of Eely. & 1 Low & 3 High & 1 High & 3\\
Power supply failure. & Eely losses its power supply & Short circuit in one of Eely's modules/sensors/thrusters. Lose cable. Water leakage. overheating of the batteries module. & Loss of Eely. & 1 Low & 3 High & 2 Med & 6\\
Propulsion system failure. & Thruster module failure & Wear of the thrusters blades. Failure of the electronic speed controller (ESC). Stucked algae and sediments over the thrusters/blades. Salt contamination over the thrusters & Loss of Eely. & 1 Low & 3 High & 1 High & 3\\
 & Joint module failure & Wear of the mechanical components of the joint module. Wear of the joint module motors. Uncoordinated movements of the joint module. Leakage of oil from the cushions insulating the joint module. & Loss of Eely. & 1 Low & 3 High & 2 Med & 6\\
 & Shape configuration failure & Uncoordinated movements of the joint module due to uncontrolled stiffness variance. Strong water pressure on the joint module. & Loss of Eely. & 1 Low & 3 High & 3 Low & 9\\
Controller failure. & Failure to supply the desired inputs to the thruster module and joint module. Failure to follow the safe distance between Eely and the seabed. & Software failure. Sensor failure. Hardware failure. & Loss of Eely. Reduced mission duration & 2 Med & 3 High & 2 Med & 12\\
 & Failure to differentiate between contouring and platforming operations & Software failure. Sensor failure. & Inaccurate mapping of the seabed. Damage to Eely or one of its modules. Loss of Eely. & 1 Low & 3 High & 1 High & 3\\
DVL failure. & Failure to accurately detect the distance between Eely and the seabed. & Software failure. Sensor failure. & Inaccurate mapping of the seabed. Damage to Eely or one of its modules. Loss of Eely. & 3 High & 3 High & 2 Med & 18\\
 & Poor detection of seabed features. & Software failure. Sensor failure. & Inaccurate mapping of the seabed. Reduced mission duration. Damage to Eely or one of its modules. Loss of Eely. & 2 Med & 1 Low & 1 High & 2\\
UHI failure. & Failure to accurately detect the distance between Eely and the seabed. & Software failure. Sensor failure. & Inaccurate mapping of the seabed. Reduced mission duration. Damage to Eely or one of its modules due to collision due to poor control decision of whether to map on a contouring or platforming manner. & 1 Low & 2 Mid & 1 High & 2\\
 & Poor detection of seabed features. & Software failure. Sensor failure. & Inaccurate mapping of the seabed. Reduced mission duration. Damage to Eely or one of its modules due to collision due to poor control decision of whether to map on a contouring or platforming manner. & 1 Low & 2 Mid & 1 High & 2
\end{tblr}
}
\label{Table:PHA seabed}
\end{table}

%///////////////
% \usepackage{tabularray}
\begin{table}
\centering
\caption{Relevant hazards for confined environments operations of Eely}
\resizebox{\linewidth}{!}{%
\begin{tblr}{
  width = \linewidth,
  colspec = {Q[70]Q[175]Q[250]Q[250]Q[50]Q[50]Q[50]Q[40]},
  row{1} = {c},
  cell{1}{1} = {r=2}{},
  cell{1}{2} = {r=2}{},
  cell{1}{3} = {r=2}{},
  cell{1}{4} = {r=2}{},
  cell{1}{5} = {c=4}{0.161\linewidth},
  cell{3}{1} = {r=3}{},
  cell{7}{1} = {r=3}{},
  cell{10}{1} = {r=2}{},
  cell{13}{1} = {r=2}{},
    vlines,
  % hline{1,3,5,7-8,11,13,15,17} = {-}{},
  % hline{2} = {5-8}{},
  % hline{4,6,9-10,12,14,16} = {2-8}{},
  hline{1,3,6-7,10,12-13,15} = {-}{},
  hline{2} = {5-8}{},
  hline{4-5,8-9,11,14} = {2-8}{},
}
\textbf{ Hazard} & \textbf{ Hazardous Event} & \textbf{ Cause} & \textbf{ Consequence} & \textbf{Risk} &  &  & \\
 &  &  &  & Freq. & Conseq. & Detect. & Prod. (rpn)\\
Strong currents. & Currents are too strong for Eely to handle. & Currents are not accounted for in the mission plan. & Reduced mission duration. Reduced maneuverability in different configurations for Eely. & 1 Low & 2 Med & 1 High & 2\\
 & Strong currents lead Eely to hit the cave or tunnel. & Eely is doing contouring for the inside of a cave or tunnel. & Damage to Eely or one of its modules. Loss of Eely. & 2 Med & 3 High & 2 Med & 12\\
 & Eely is stuck in soft sediment and algae in cave or tunnel. & Failure to detect the cave or tunnel. Failure to follow the safe distance between Eely and the inside of cave or tunnel. Loss of Eely & Damage to Eely or one of its modules. Loss of Eely. & 2 Med & 3 High & 3 Low & 18\\
Power supply failure. & Eely losses its power supply & Short circuit in one of Eely's modules/sensors/thrusters. Lose cable. Water leakage. overheating of the batteries module. & Loss of Eely. & 1 Low & 3 High & 2 Med & 6\\
Propulsion system failure. & Thruster module failure & Wear of the thrusters blades. Failure of the electronic speed controller (ESC). Stucked algae and sediments over the thrusters/blades. Salt contamination over the thrusters. High water pressure. & Loss of Eely. & 1 Low & 3 High & 2 Med & 6\\
 & Joint module failure & Wear of the mechanical components of the joint module. Wear of the joint module motors. Uncoordinated movements of the joint module. High water pressure. Leakage of oil from the cushions insulating the joint module. & Loss of Eely. & 1 Low & 3 High & 2 Med & 6\\
 & Shape configuration failure & Uncoordinated movements of the joint module due to uncontrolled stiffness variance. Strong water pressure on the joint module. & Loss of Eely. & 1 Low & 3 High & 3 Low & 9\\
Controller failure. & Failure to supply the desired inputs to the thruster module and joint module. Failure to follow the safe distance between Eely and the seabed. & Software failure. Sensor failure. Hardware failure. & Loss of Eely. Reduced mission duration & 2 Med & 3 High & 2 Med & 12\\
 & Failure to differentiate between contouring and platforming operations & Software failure. Sensor failure. & Inaccurate mapping of the inside of cave or tunnel. Damage to Eely or one of its modules. Loss of Eely. & 1 Low & 3 High & 1 High & 3\\
DVL failure. & Failure to accurately detect the distance between Eely and the inside of cave or tunnel. & Software failure. Sensor failure. & Inaccurate mapping of the inside of cave or tunnel. Damage to Eely or one of its modules. Loss of Eely. & 3 High & 3 High & 2 Med & 18\\
UHI failure. & Failure to accurately detect the distance between Eely and the inside of cave or tunnel. & Software failure. Sensor failure. & Inaccurate mapping of the the inside of cave or tunnel. Reduced mission duration. Damage to Eely or one of its modules due to collision. & 1 Low & 2 Med & 1 High & 2\\
 & Poor detection of the inside of cave or tunnel features. & Software failure. Sensor failure. & Inaccurate mapping of the the inside of cave or tunnel. Reduced mission duration. Damage to Eely or one of its modules due to collision. & 1 Low & 2 Med & 1 High & 2
\end{tblr}
}
\label{Table:PHA confined}
\end{table}
%///////////////

\end{document}